\def\paperlanguage{}
\pdfoutput=1 % for arxiv

\documentclass[letterpaper, 10 pt, conference]{ieeeconf}  % Comment this line out if you need a4paper

\usepackage{bm}
\usepackage{cite}
\usepackage{flushend}
\include{preamble}

\IEEEoverridecommandlockouts                              % This command is only needed if
\title{\LARGE \textbf
  {
    \switchlanguage%
    {%
      TWIMP: Two-Wheel Inverted Musculoskeletal Pendulum\\ as a Learning Control Platform\\ in the Real World with Environmental Physical Contact
    }%
    {%
      環境接触を伴う学習型制御開発のプラットフォームを目指した\\筋骨格型倒立二輪TWIMP
    }%
  }
}

\author{Kento Kawaharazuka$^{*,1}$, Tasuku Makabe$^{*,1}$, Shogo Makino$^{1}$, Kei Tsuzuki$^{1}$, Yuya Nagamatsu$^{1}$, Yuki Asano$^{1}$\\Takuma Shirai$^{1}$, Fumihito Sugai$^{1}$, Kei Okada$^{1}$, Koji Kawasaki$^{2}$ and Masayuki Inaba$^{1}$% <-this % stops a space
  \thanks{$^{*}$ K. Kawaharazuka and T. Makabe contributed equally to this work.}
  \thanks{$^{1}$ The authors are with the Department of Mechano-Informatics, Graduate School of Information Science and Technology, The University of Tokyo, 7-3-1 Hongo, Bunkyo-ku, Tokyo, 113-8656, Japan.
    {\texttt\small [kawaharazuka, makabe, makino, tsuzuki, nagamatsu, asano, t\_shirai, sugai, k-okada, inaba]@jsk.t.u-tokyo.ac.jp}
  }
  \thanks{$^{2}$ The author is associated with TOYOTA MOTOR CORPORATION.
    {\texttt\small koji\_kawasaki@mail.toyota.co.jp}
  }
}
\begin{document}

\maketitle
\thispagestyle{empty}
\pagestyle{empty}

%%%%%%%%%%%%%%%%%%%%%%%%%%%%%%%%%%%%%%%%%%%%%%%%%%%%%%%%%%%%%%%%%%%%%%%%%%%%%%%%
\begin{abstract}
  \switchlanguage%
  {%
    By the recent spread of machine learning in the robotics field, a humanoid that can act, perceive, and learn in the real world through contact with the environment needs to be developed.
    In this study, as one of the choices, we propose a novel humanoid TWIMP, which combines a human mimetic musculoskeletal upper limb with a two-wheel inverted pendulum.
    By combining the benefit of a musculoskeletal humanoid, which can achieve soft contact with the external environment, and the benefit of a two-wheel inverted pendulum with a small footprint and high mobility, we can easily investigate learning control systems in environments with contact and sudden impact.
    We reveal our whole concept and system details of TWIMP, and execute several preliminary experiments to show its potential ability.
  }%
  {%
    近年の機械学習の普及によって, より環境と接触し, 実世界で学習型制御を行うことのできるヒューマノイドが求められている.
    本研究では, その一つの選択肢として, 人体を模倣した筋構造を有する筋骨格ヒューマノイドと, 倒立二輪を組み合わせた新しいヒューマノイドTWIMPを提案する.
    筋骨格ヒューマノイドの持つ外界との柔らかな接触や可変剛性制御等の利点と, 倒立二輪のフットプリントが小さく移動性能の高いという二つの点のいいとこ取りをし, それらを組み合わせることで, より接触や衝撃の加わる環境における学習型制御の模索を容易とする.
    全体の設計コンセプトとシステムの詳細を明らかにし, そのポテンシャルを示すため, いくつかのpreliminaryな実験を行う.
  }%
\end{abstract}

%%%%%%%%%%%%%%%%%%%%%%%%%%%%%%%%%%%%%%%%%%%%%%%%%%%%%%%%%%%%%%%%%%%%%%%%%%%%%%%%
\section{INTRODUCTION} \label{sec:introduction}
\switchlanguage%
{%
  In recent years, the spread of machine learning methods such as deep learning is rapid, and there have been big achievements in fields of image classification \cite{krizhevsky2012imagenet}, sentence generation \cite{karpathy2015deep}, etc.
  Also, machine learning is spreading in the robotics field, such as \cite{zhang2015towards, tobin2017domain, levine2018learning}, and a humanoid that can act, perceive, and learn in the real world through contact with the environment needs to be developed.

  Therefore, we propose a two-wheel inverted musculoskeletal pendulum TWIMP, as one of the choices to investigate learning control systems in the actual environment with contact and sudden impact.
  Our concept is a simple one that uses a tendon-driven musculoskeletal structure as an upper limb, and uses a two-wheel inverted pendulum as a lower limb.
  Because the musculoskeletal humanoid \cite{mizuuchi2016advanced, ijars2013:nakanishi:approach, artl2013:wittmeier:ecce, humanoids2016:asano:kengoro} imitates human structures, it has many benefits such as the flexible spine, underactuated fingers, error recovery using redundant muscles, variable stiffness control using antagonism and nonlinear elastic feature of muscles, and joint torque control using muscle tension.
  Also, because the two-wheel inverted pendulum \cite{ijrs2005:kin:pendulum, iros2007:jeong:pendulum} has high mobility and a small footprint, it can enter cluttered narrow spaces.
  By combining the benefits of these two, we can make learning control in the real world easier.
  In this study, we develop the two-wheel inverted musculoskeletal pendulum TWIMP, and conduct several preliminary experiments to show its potential ability.
}%
{%
  近年, 深層学習等の機械学習手法の普及はめまぐるしく, 画像分類\cite{krizhevsky2012imagenet}や文章生成\cite{karpathy2015deep}等の分野において大きな功績を残している.
  また, ロボットの動作生成等にも大きな躍進を見せており\cite{zhang2015towards, tobin2017domain, levine2018learning}, より実世界で実験を行っていくロボットが求められてきている.

  そこで我々は, 実機における環境接触等を含む学習型制御模索のための一つの選択肢として, 筋骨格型倒立二輪TWIMPを提案する.
  これは上肢として筋骨格ヒューマノイドを用い, 下肢として倒立二輪を用いる, という単純なコンセプトである.
  筋骨格ヒューマノイド\cite{ijars2013:nakanishi:approach, artl2013:wittmeier:ecce, humanoids2013:michael:anthrob, humanoids2016:asano:kengoro}は, 背骨等の多椎構造, 指等の劣駆動構造, 冗長な筋配置によるエラーリカバリ, 筋の拮抗構造と非線形弾性要素を利用したハードウェア可変剛性制御, 筋張力による関節トルク制御等, 人体を模倣しているがゆえの利点を多く有する.
  その柔軟な身体は, 環境との接触を含むような行動生成に相応しく, 衝撃に対するリカバリや筋の冗長性による継続した学習に対しても応用可能である.
  また, 二輪型の倒立振子\cite{ijrs2005:kin:pendulum, iros2007:jeong:pendulum}は移動性能に優れ, 足のフットプリントも小さいため, 狭隘な環境にも入って行くことができると考える.
  この二つの利点を組み合わせることで, より実世界における学習型制御に対するハードルを低くすることができると考え, 本研究ではそのための筋骨格型倒立二輪TWIMPの開発と, このヒューマノイドを用いたpreliminaryないくつかの実験を行う.
}%

\begin{figure}[t]
  \centering
  \includegraphics[width=0.9\columnwidth]{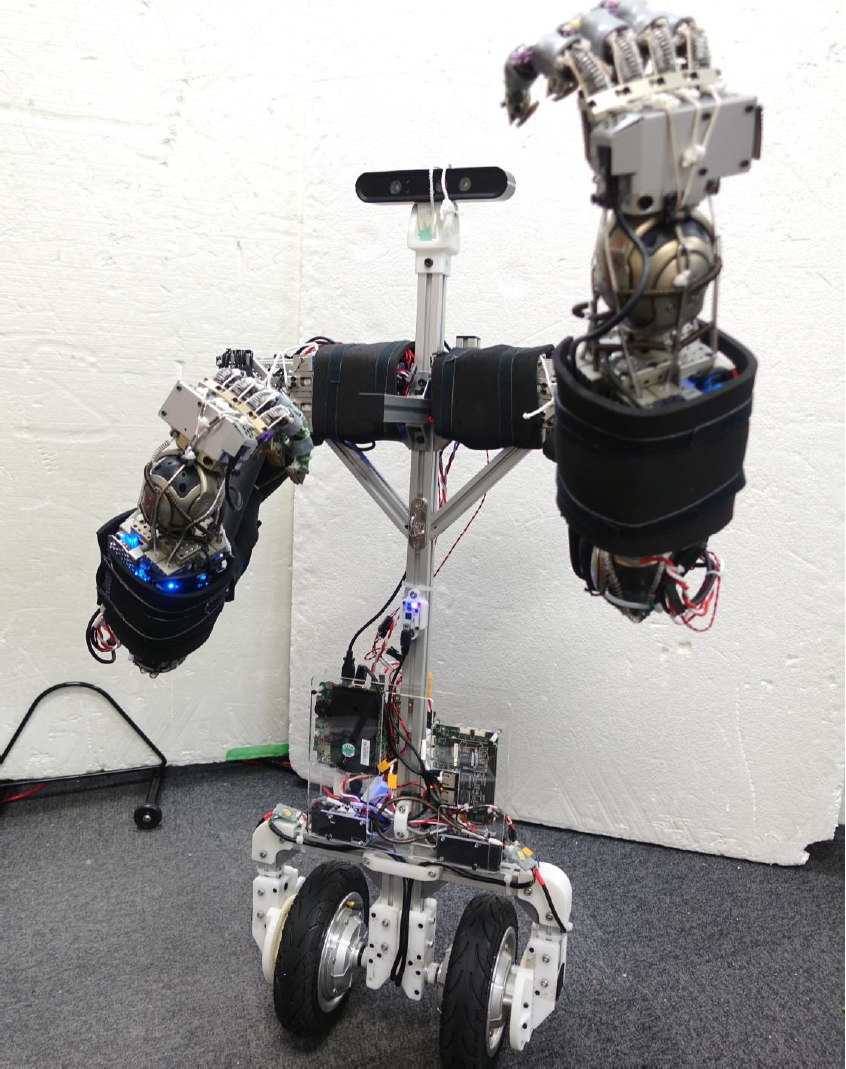}
  \caption{Overview of TWIMP: Two-Wheel Inverted Musculoskeletal Pendulum}
  \label{figure:twimp-overview}
  \vspace{-3.0ex}
\end{figure}

%%%%%%%%%%%%%%%%%%%%%%%%%%%%%%%%%%%%%%%%%%%%%%%%%%%%%%%%%%%%%%%%%%%%%%%%%%%%%%%%
\section{Our Approach: Musculoskeletal Humanoid with a Two-wheel Inverted Pendulum} \label{sec:approach}
\subsection{Overview of Our Proposed Approach}
\switchlanguage%
{%
  We propose a hybrid humanoid which has a musculoskeletal structure in the upper limb and a two-wheel inverted pendulum in the lower limb.
  The components of the musculoskeletal upper limb are fully modularized, and it can be constructed and reconstructed easily.
  The structure of the two-wheel inverted pendulum is constructed using a generic aluminum frame as with the musculoskeletal upper limb, and the modification of the link lengths and rearrangement of links are easy.
  By combining the two, we construct a two-wheel inverted musculoskeletal pendulum TWIMP, which has high impact resistance due to the flexible body structure and high mobility, and which can be reconstructed according to the needs.

  We control TWIMP using the general state feedback method with the optimal regulator, and add a control scheme to modify the upper body posture depending on the movements of the upper limb.
  We control the musculoskeletal upper limb using self-body image \cite{ral2018:kawaharazuka:vision-learning, iros2018:kawaharazuka:bodyimage} and joint torque control considering muscle tension \cite{humanoids2016:kawamura:controll}.

  We conduct basic movement, manipulation, and impact resistance experiments to show the potential ability of TWIMP, and discuss its control problems and its next generation design with a spine and tail.
}%
{%
  上半身は筋骨格ヒューマノイド, 下半身は倒立二輪によるハイブリッド構成とする.
  筋骨格上肢は全身のコンポーネントがモジュール化されており, 簡易に構成・そして再構成可能な筋骨格ヒューマノイドとする.
  倒立二輪はIn-wheel motorを使った一般的な車輪構成とし,　筋骨格上肢と同様汎用のアルミフレームで構成できるようにし, リンク長の変更や組み換えを容易な構造とする.
  この二つを合わせることによて, ニーズに合わせて再構成可能で, 柔軟な身体構造による対衝撃性を有し, 移動性能に優れた筋骨格型倒立二輪TWIMPを構成する.

  制御は一般的な状態フィードバックにおける最適レギュレータを用い, また, 上肢の動作に応じて姿勢を変化させる制御機構を取り入れる.
  筋骨格上肢は自己身体像\cite{iros2018:kawaharazuka:bodyimage}を用いた制御手法と筋張力による関節トルク制御\cite{humanoids2016:kawamura:controll}を用いる.

  このTWIMPによって移動実験・マニピュレーション実験・衝撃実験を行い, そのポテンシャルを示すと同時に, そこで出た問題について議論し, 学習型制御・背骨や尻尾を有する次世代のTWIMPについても議論を行う.
}%

\subsection{Related Works and Our Contribution}
\switchlanguage%
{%
  We will discuss the related works regarding musculoskeletal humanoids, two-wheel inverted pendulums, and robots combining wheels and manipulators, separately.

  So far, tendon-driven musculoskeletal humanoids such as Kenshiro \cite{ijars2013:nakanishi:approach} and Kengoro \cite{humanoids2016:asano:kengoro} have been developed.
  These robots aimed to mimic the musculoskeletal structures of human beings, and have underactuated structures, flexible muscle structures, etc.
  The controls of the upper limb such as variable stiffness control and muscle tension control have been developed to a practical level.
  However, regarding the movements of the lower limb, although controls such as the balancing control using muscle-ZMP \cite{asano2016human} exist, precise control of the soft lower limb that supports the upper limb and walks stably is difficult, and the bipedal locomotion of the musculoskeletal humanoid has not been achieved yet.
  Therefore, manipulation with locomotion using the musculoskeletal humanoid has rarely been studied.

  Regarding the two-wheel inverted pendulum, there are many studies such as \cite{ijrs2005:kin:pendulum} and \cite{iros2007:jeong:pendulum}, and the mobility is high.
  Also, it can move stably even on slopes.
  At the same time, trolly type robots can move more stably on level ground and no force is needed to stabilize the posture.
  Also, there are problems that both of them generally cannot get up after falling down.

  Popular trolly type robots that combine wheels and manipulators are PR2 \cite{PR2:WillowGarage}, Fetch \cite{fetch:FetchRobotics}, etc., and these play an active part in the robotics research field.
  However, because these robots cannot enter cluttered narrow spaces due to their big footprints, and cannot move stably on slopes, robots with not trollies but with wheeled inverted pendulums for locomotion have been developed.
  Handle of Boston Dynamics \cite{bostondynamics:handle} has a wheel at each tip of two legs that move independantly, and can move stably using the multiple degrees of freedom on hilly landscapes that rise and fall.
  Emiew of HITACHI \cite{iros2006:hosoda:emiew} has a human-like upper limb with dual arms and a head, and has a two-wheel inverted pendulum with high mobility, and it was popular as a human friendly navigation robot.
  However, all of these robots have rarely moved in situations with environmental physical contact, and they were developed with the weight on the mobility of the two-wheel inverted pendulum.
  Although UBot-5 \cite{kuindersma2013pendulum} has two arms and a wheeled inverted pendulum in its small body and does learning control with environmental contact, we need to develop a life-sized humanoid that can be used more practically in the real world such as livelihood support and has more flexible arms with the ability to endure impacts due to large inertia.

  Our contribution is that we propose a robot combining the musculoskeletal structure, which has high performance regarding soft contact with the environment and impact resistance, with the two-wheel inverted pendulum, which has a small footprint and high mobility, to be able to neutralize the disadvantages of these two and emphasize their benefits.
  This robot may be able to get up using the two wheels and flexible musculoskeletal dual arm, after falling down.
  Also, as a research platform of learning control systems, an easy design which can be constructed and reconstructed is desired, and this proposed robot is effective because it is designed using generic aluminum frames for almost all of the structures.
}%
{%
  これまでの先行研究を, 筋骨格ヒューマノイド, 倒立二輪, 車輪とマニピュレータを組み合わせたタイプのロボットに分けて議論する.

  筋骨格ヒューマノイドは今まで, Kenshiro\cite{ijars2013:nakanishi:approach}やKengoro\cite{humanoids2016:asano:kengoro}等が開発されている.
  これらは詳細な人体模倣を目指しており, 人体における劣駆動性や柔軟性等を目指したものである.
  同時に, 上肢の可変剛性制御や筋張力制御等は実用的なレベルまで模索されてきたのに対して, 下肢の動きに関しては, 筋張力から計算したZMPによるバランス制御\cite{asano2016human}等行われているが, 上半身を支えて安定して歩行するための柔軟な身体の精密な制御が困難であるため, 未だに二脚での歩行による移動は達成されていない.
  そのため, 筋骨格ヒューマノイドは移動しながらのマニピュレーションについてはほとんど研究されることはなかった.

  倒立二輪としては\cite{ijrs2005:kin:pendulum}や\cite{iros2007:jeong:pendulum}のようなものが挙げられ, これまで盛んに研究されてきており, 移動性や安定性に関して優れていると言える.
  同時に, 台車型のロボットはより安定しており, 姿勢安定化に力を要さないという利点がある.
  また, どちらもそのままでは基本的には倒れたら起き上がることはできないという問題がある.

  車輪とマニピュレータを組み合わせたタイプのロボットとして代表的なのはPR2\cite{PR2:WillowGarage}やFetch\cite{fetch:FetchRobotics}等のロボットであり, 研究用として, 多くの場所において活躍している.
  しかし, これらのロボットは足のフットプリントが大きいため狭隘な環境で作業することは難しく, 一度倒れると終わりなため, 足を台車ではなく倒立二輪としたロボットも多く開発されてきている.
  Boston DynamicsのHandle\cite{bostondynamics:handle}は独立に可動する脚の先端に車輪を持ち、自由度を用いて起伏のある地面でも常に両輪を接地させつつ安定した走行を行う。
  HITACHIのEmiew\cite{iros2006:hosoda:emiew}は双腕と頭部を持つ人間に近い上半身と、移動性能に優れる倒立振子を組み合わせて、人間にとって親しみやすいナビゲーションロボットとして活躍した。
  しかし, どれもマニピュレータを環境と接触しながら動作することは少なく, 倒立二輪の走行性能を重要視したロボットが多く開発されてきた.
  UBot-5 \cite{kuindersma2013pendulum}は小さな体に二本の腕と倒立二輪を持ち, 環境接触を伴いながらの学習型制御を行っているが, より日常生活等において実用的に用いることのできる大きさかつ, その大きな慣性による衝撃を受け止めることのできる柔らかな腕が必要である.

  我々のContributionは, さらにマニピュレータを環境と接触, また強い衝撃を加えるような動作に対しての利用を考え, それら性能に優れた筋骨格構造を持つ上肢と移動性能・フットプリントの小さな倒立二輪を組み合わることでお互いの欠点を相殺し, 利点をより強調することのできるロボットを提案したところにある.
  また, 学習等における研究用プラットフォームとしては, 簡易に構成・再構成可能な設計が望ましく, その点でもその骨格のほとんどに汎用フレームを用いた本提案は有効であると考える.
}%

%%%%%%%%%%%%%%%%%%%%%%%%%%%%%%%%%%%%%%%%%%%%%%%%%%%%%%%%%%%%%%%%%%%%%%%%%%%%%%%%
\section{Design of TWIMP} \label{sec:design}
\switchlanguage%
{%
  In this section, we show the whole design of TWIMP.
  First, we explain the musculoskeletal dual arm and two-wheel inverted pendulum separately, and then explain TWIMP composed of these two.
}%
{%
  本章では, TWIMPの全体設計を示す.
  初めに筋骨格双腕, 倒立二輪について分けて説明し, 最後にそれを合体させた筋骨格倒立二輪について説明する.
}%

\subsection{Tendon-driven Musculoskeletal Dual Arm} \label{subsec:upperlimb-design}
\begin{figure}[htb]
  \centering
  \includegraphics[width=0.9\columnwidth]{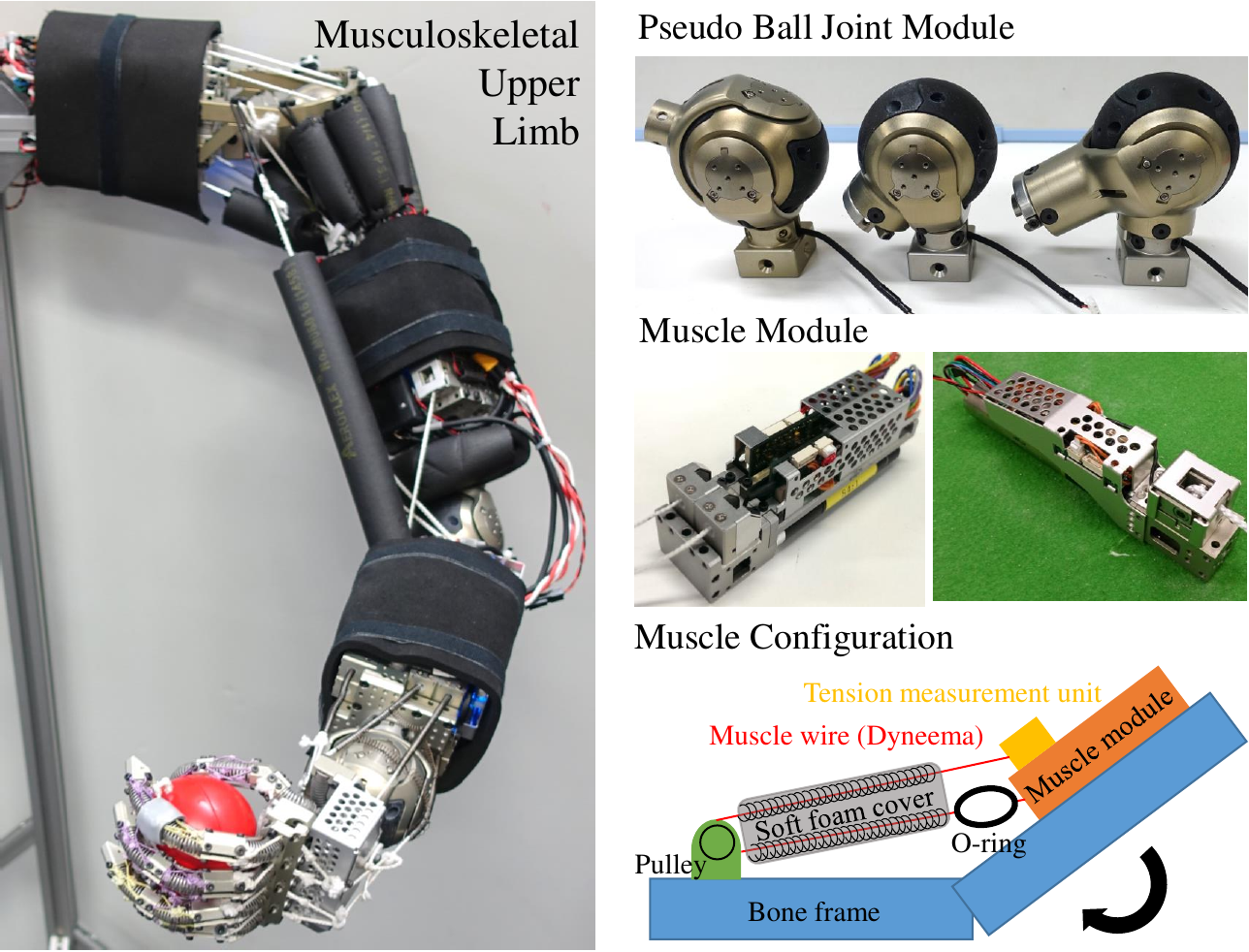}
  \caption{Design of musculoskeletal upper limb}
  \label{figure:upperlimb-design}
  \vspace{-1.0ex}
\end{figure}

\switchlanguage%
{%
  The musculoskeletal dual arm is an extension of the single arm proposed in \cite{robomech2018:kawaharazuka:musashilarm}.
  We show the detailed design in \figref{figure:upperlimb-design}.
  This arm has an easily reconfigurable structure composed of joint module, muscle module, and bone frame.
  Since a generic aluminum frame is used as the bone structure, we can construct the body structure freely and easily by combining the frames using brackets.
  The muscle module contains several sensors, which measure motor temperature, muscle length, and muscle tension.
  Since each joint module contains an IMU and potentiometers, we can obtain sensor information redundantly.

  In addition, since the muscle is composed of a chemical fiber Dyneema and O-ring used as a nonlinear elastic element, it has a soft hardware structure.
  The nonlinear elastic element and antagonistic structure of the musculoskeletal humanoid enable variable stiffness control of the body joint.
}%
{%
  本研究で用いる筋骨格双腕は, \cite{robomech2018:kawaharazuka:musashilarm}で提案されたものを双腕に拡張したものである.
  設計の詳細を\figref{figure:upperlimb-design}に示す.
  これは全身の関節, 筋, 骨等を全てモジュール化することで, 簡易に構成・そして再構成可能な筋骨格ヒューマノイドとなっている.
  骨格は汎用的なアルミ製汎用フレームのを用いており, ブラケットを使うことで自由に汎用フレームを組み合わせ, 骨格を形成することができる.
  筋モジュールにはセンサとして筋の温度を測るセンサ, 筋の長さを測るセンサ, 筋の張力を測るセンサが入っている.
  関節モジュールにはそれぞれIMUとPotentiometerが入っているため, 冗長にセンサ情報を取得できるようになっている.

  また, 筋は化学繊維であるダイニーマと非線形弾性要素としてのOリングを用いたハードウェアとして柔らかい構造となっている.
  非線形弾性特性と, 筋骨格腱駆動ヒューマノイド特有の拮抗構造を用いることで, 身体の剛性を可変にすることができるという特徴を有する.
}%

\subsection{Two-wheel Inverted Pendulum} \label{subsec:pendulum-design}
\begin{figure}[htb]
  \centering
  \includegraphics[width=0.8\columnwidth]{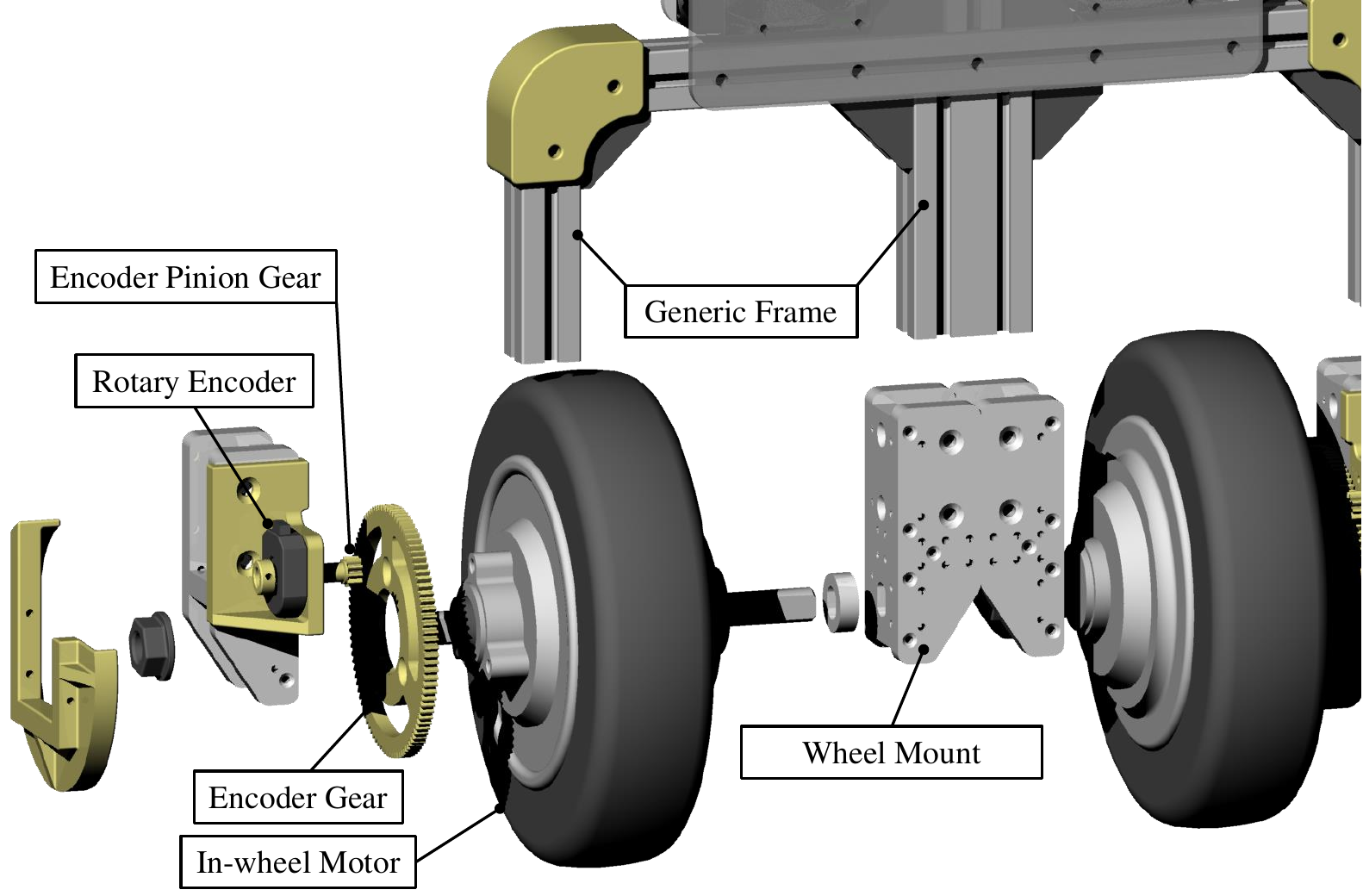}
  \caption{Design of two-wheel inverted pendulum}
  \label{figure:pendulum-design}
  \vspace{-1.0ex}
\end{figure}

\switchlanguage%
{%
  The detailed design of the two-wheel inverted pendulum is shown in \figref{figure:pendulum-design}.
  We chose the wheel according to its output torque and easily configurable structure.
  So we used an in-wheel motor unit composed of a motor and wheel, and a generic aluminum frame as the bone structure, and achieved an easily reconfigurable simple wheel structure.
  TWIMP has rotary encoders outside of the in-wheel motor, and IMU at the trunk frame, and we control the posture, transition, and rotation of the inverted pendulum using the information from these sensors.
}%
{%
  倒立二輪の設計詳細を\figref{figure:pendulum-design}に示す.
  車輪はトルクを重視することと, なるべく簡易な構成にすることで, 再構成を容易にしている.
  そこで, モータと車輪が一体となったインホイールモータを使用しており, 外側にアタッチメントをつけ, 骨格は汎用フレームで構成することで非常に簡易な構成にすることに成功している.
  インホイールモータには外付けでエンコーダが, 体幹の骨格にはIMUが載っており, この二つを用いて倒立振子の姿勢・位置・角度に関する制御を行っている.
}%

\subsection{Two-wheel Inverted Musculoskeletal Pendulum} \label{subsec:twimp-design}
\begin{figure}[htb]
  \centering
  \includegraphics[width=0.95\columnwidth]{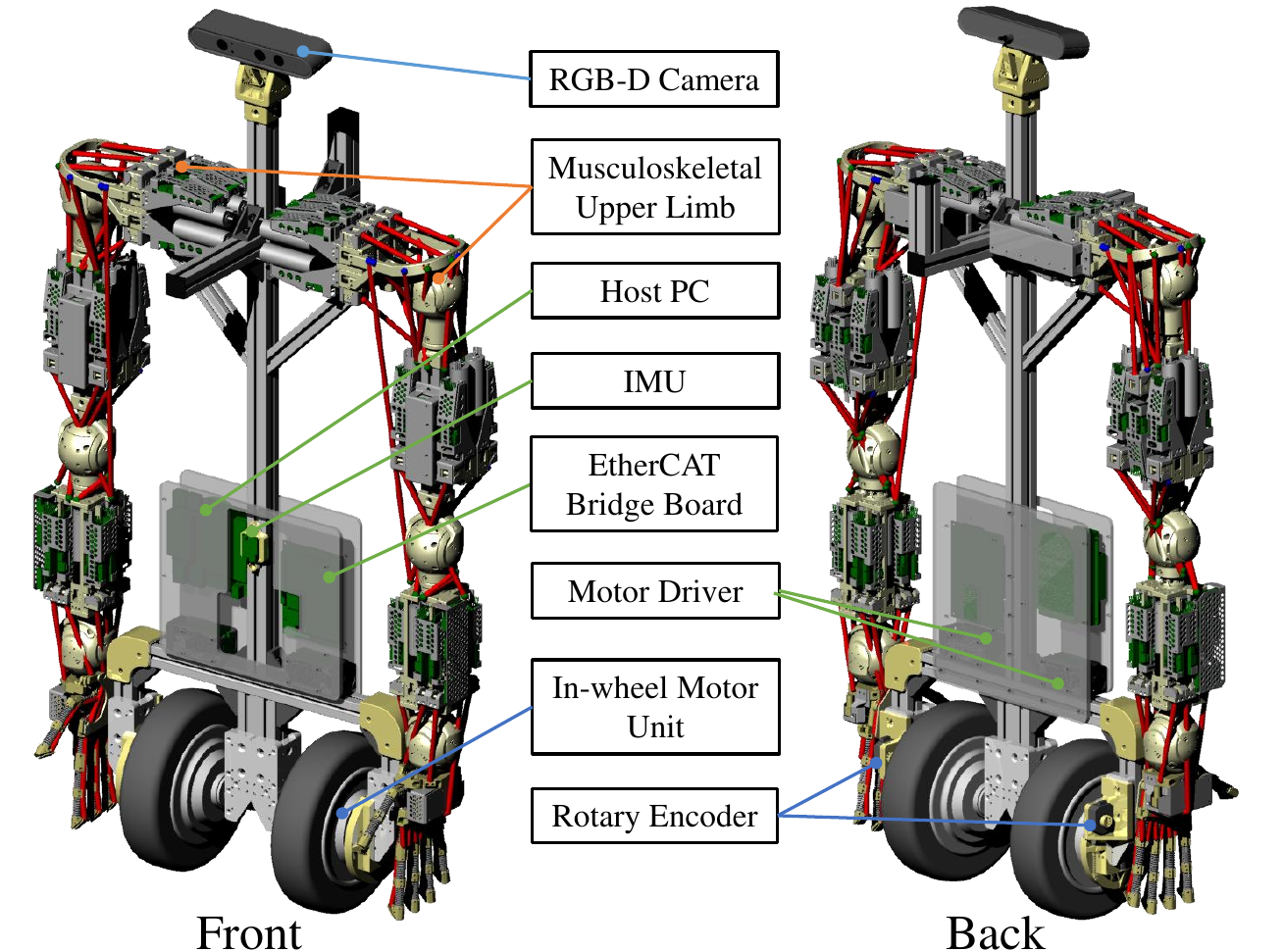}
  \caption{Design of TWIMP}
  \label{figure:twimp-design}
  \vspace{-1.0ex}
\end{figure}

\begin{figure}[htb]
  \centering
  \includegraphics[width=0.95\columnwidth]{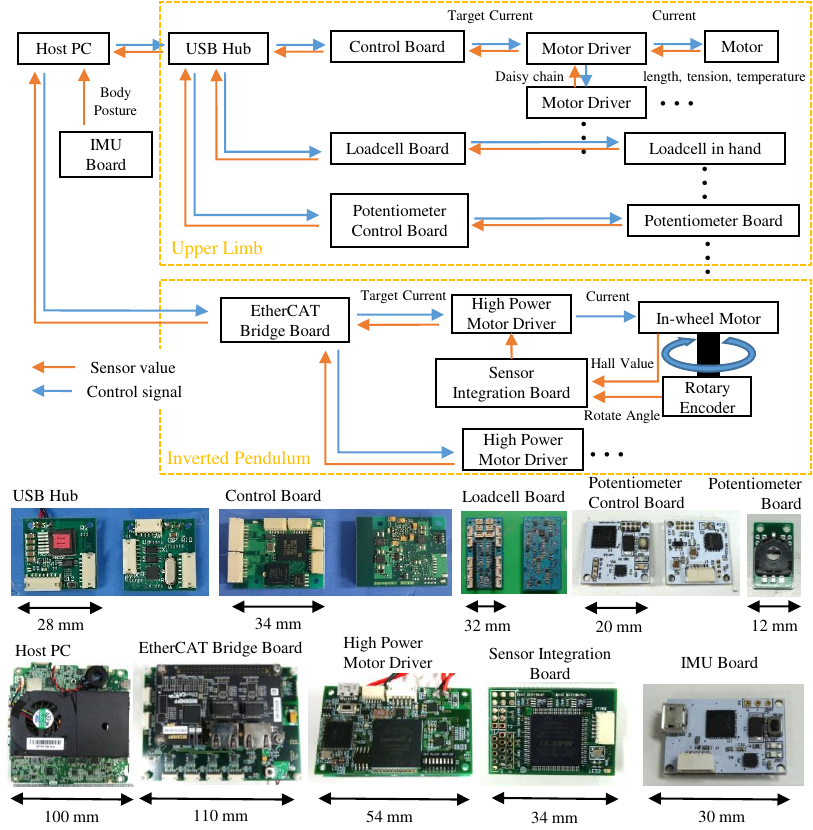}
  \caption{Circuit configuration of TWIMP}
  \label{figure:twimp-circuit}
  \vspace{-1.0ex}
\end{figure}

\switchlanguage%
{%
  The detailed design of TWIMP is shown in \figref{figure:twimp-design}.
  Since TWIMP is the combination of the upper limb in \secref{subsec:upperlimb-design} and the lower limb in \secref{subsec:pendulum-design} using generic frames, we can change the link lengths easily.
  TWIMP has an AstraS camera as a RGB-D vision sensor on its head.
  The circuit configuration is shown in \figref{figure:twimp-circuit}.
  The circuit system of the high power humanoid JAXON \cite{kojima2015development} is used for the lower limb, and that of the musculoskeletal humanoid \cite{robomech2018:kawaharazuka:musashilarm}, with a space-saving characteristic, is used for the upper limb.
  To avoid damaging the circuits by its contact with the ground when TWIMP falls down, it has generic frames in the front and back of its chest frame.
}%
{%
  全体の構成を\figref{figure:twimp-design}に示す.
  基本的には, \secref{subsec:upperlimb-design}と\secref{subsec:pendulum-design}を汎用フレームにより合体させた身体となっており, 自由にリンク長を変えることができる.
  頭にはRGBD視覚としてAstraSがついている.
  回路構成は\figref{figure:twimp-circuit}のようになっている.
  JAXON\cite{humanoids2015:kojima:jaxon}における大出力型ロボットの回路構成を足回りに対して適用し, 筋骨格ヒューマノイド\cite{robomech2018:kawaharazuka:musashilarm}における省スペースを重視した回路構成を組み合わせた構成となっており, HostPC(Intel NUC)やIMU等は体幹部分に収納されている.
  転倒時に回路が破損することを防ぐため、上肢の胸部の前後にフレームを備え、転倒時に回路が直接地面と接触することを防いでいる。
}%

%%%%%%%%%%%%%%%%%%%%%%%%%%%%%%%%%%%%%%%%%%%%%%%%%%%%%%%%%%%%%%%%%%%%%%%%%%%%%%%%
\section{Control of TWIMP} \label{sec:control}
\switchlanguage%
{%
  In this section, we will explain the control system of TWIMP.
  First, we will explain the musculoskeletal dual arm and two-wheel inverted pendulum separately, and then, explain the integrated control system of TWIMP.
}%
{%
  本章では, TWIMPの制御システムについて説明する.
  初めに筋骨格双腕, 倒立二輪について分けて説明し, 最後にそれを合体させた筋骨格倒立二輪の制御手法について説明する.
}%

\subsection{Control of Musculoskeletal Dual Arm} \label{subsec:upperlimb-control}
\switchlanguage%
{%
  As the control of the musculoskeletal dual arm, (1) the control using the self-body image acquired with online training \cite{iros2018:kawaharazuka:bodyimage} and (2) the torque control using muscle tension \cite{humanoids2016:kawamura:controll} are possible.
  Concerning (1), we define the self-body image as below.
  \begin{align}
    \bm{l}_{target} = \bm{f}_{ideal}(\bm{\xi}_{target}) + \bm{g}(\bm{\xi}_{target}, \bm{T}) \label{eq:self-body-image}
  \end{align}
  $\bm{l}_{target}$ is the target muscle lengths, $\bm{f}_{ideal}$ is the joint-muscle mapping in an ideal situation without external force, $\bm{\xi}_{target}$ is the target joint angles, $\bm{g}$ is the compensation term of the extension of muscles, and $\bm{T}$ is the current muscle tensions.
  In \cite{iros2018:kawaharazuka:bodyimage}, this mapping is acquired by the initial training using the geometric model and online training using the actual robot.
  In this study, we trained the self-body image in advance, and conducted basic position control using \equref{eq:self-body-image}.

  Concerning (2), the target joint torques are calculated by proportional control, and the target muscle tensions are calculated to minimize the sum of the squares of muscle tensions as shown below.
  \begin{align}
    &\bm{\tau}_{ref} = K_j(\bm{\xi}_{ref}-\bm{\xi}) + \bm{\tau}_g(\bm{\xi})\\
    &\textrm{minimize}\ \ \ \bm{x}^TW\bm{x}\\
    &\textrm{subject to}\ \ \ \bm{\tau}_{ref} = -G^T\bm{x}\\
    &\ \ \ \ \ \ \ \ \ \ \ \ \ \ \ \ \ \bm{x} \geq \bm{T}_{min}
  \end{align}

  $\bm{\tau}_{ref}$ is the reference of joint torques, $K_j$ is the gain of the proportional control, $\bm{\xi}_{ref}$ is the reference of joint angles, $\bm{\xi}$ is the current joint angles, $\bm{\tau}_g$ is the torque of gravity compensation, $\bm{x}$ is the reference of muscle tensions ($T_{ref}$), $W$ is the weight matrix, $G$ is the muscle jacobian, and $\bm{T}_{min}$ is the minimum muscle tension.

}%
{%
  筋骨格上肢の制御としては, (1)オンライン学習された自己身体像を用いた制御\cite{iros2018:kawaharazuka:bodyimage}と, (2)筋張力による関節トルク制御\cite{humanoids2016:kawamura:controll}が可能となっている.
  (1)においては, 筋骨格構造における自己身体像を以下のように定義し, これを用いて動作を行う.
  \begin{align}
    \bm{l}_{target} = \bm{f}_{ideal}(\bm{\theta}_{target}) + \bm{g}(\bm{\theta}_{target}, \bm{T}) \label{eq:self-body-image}
  \end{align}
  ここで, $\bm{l}_{target}$は指令筋張, $\bm{f}_{ideal}$は外力のない理想的な状態における関節-筋空間マッピング, $\bm{\theta}_{target}$は指令関節角度, $\bm{g}$は筋の伸び等を考慮した補償項, $\bm{T}$は現在の筋張力を表す.
  \cite{iros2018:kawaharazuka:bodyimage}ではこれを幾何モデルを用いた初期学習と実機におけるオンライン学習により獲得する.
  本研究では事前にこれを学習させ, \equref{eq:self-body-image}を用いて基本的な位置制御を行う.

  また, (2)においては, 関節の指令値方向に比例制御で目標トルクを決め, 筋張力の二乗和を最小化するような筋張力の目標値を生成し, 指令する.
  具体的な式は以下のようになる.
  \begin{align}
    &\bm{\tau}_{ref} = K_j(\bm{\theta}_{ref}-\bm{\theta}) + \bm{\tau}_g(\bm{\theta})\\
    &\textrm{minimize}\ \ \ \bm{x}^TW\bm{x}\\
    &\textrm{subject to}\ \ \ \bm{\tau}_{ref} = -G^T\bm{x}\\
    &\ \ \ \ \ \ \ \ \ \ \ \ \ \ \ \ \ \bm{x} \geq \bm{T}_{min}
  \end{align}
  ここで, $\bm{\tau}_{ref}$は関節トルクの指令値, $K_j$は比例制御のゲイン, $\bm{\theta}_{ref}$は関節角度の指令値, $\bm{\theta}$は現在の関節角度, $\bm{\tau}_g$は重量補償トルク, $\bm{x}$は筋張力の指令値$\bm{T}_{ref}$, $W$は重み行列, $G$は筋長ヤコビアン, $\bm{T}_{min}$は筋張力の最小値を表す.
}%

\subsection{Control of Two-wheel Inverted Pendulum} \label{subsec:pendulum-control}
\begin{figure}[htb]
  \centering
  \includegraphics[width=0.6\columnwidth]{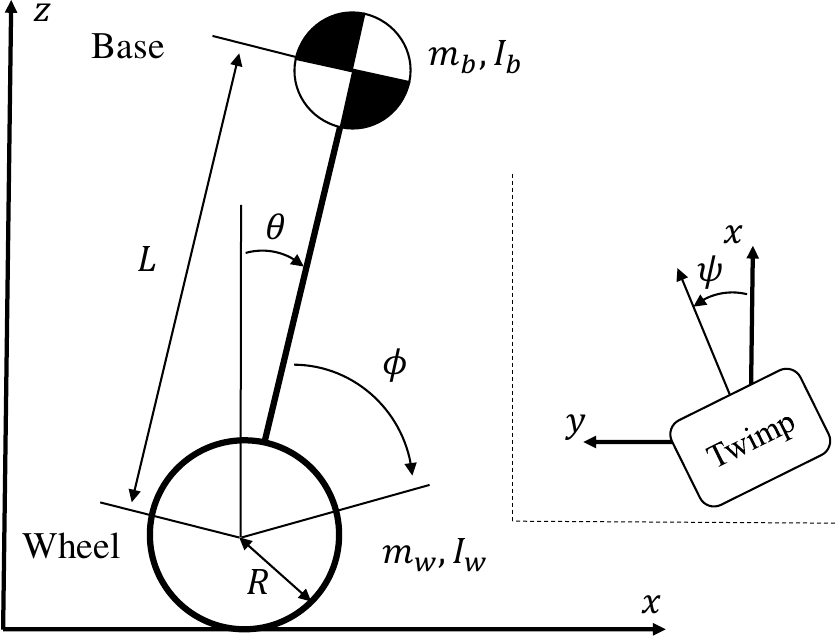}
  \caption{Parameter configuration of two-wheel inverted pendulum}
  \label{figure:pendulum-control}
  \vspace{-1.0ex}
\end{figure}

\switchlanguage%
{%
  We use the general state feedback method as the control of the two-wheel inverted pendulum.
As shown in \figref{figure:pendulum-control}, we defined that $\theta$ is the slope of the base, $\phi$ is the rotation of the wheels from the base, $m_w$ is the weight of the wheels, $m_b$ is the weight of the base, $I_w$ is the inertia of the wheels, $I_b$ is the inertia of the base, $R$ is the radius of the wheels, $L$ is the distance between the center of mass of the base and wheels, $\tau$ is the torque of the wheels, and $\bm{x}=\begin{bmatrix}\theta&\phi&\dot{\theta}&\dot{\phi}\end{bmatrix}^T$ is the state variable.
  Solving the lagrange equation, the state equation is shown as below.

  \begin{align}
    &E\dot{\bm{x}} = A_0\bm{x} + B_0\tau\\
    &\dot{\bm{x}} = A\bm{x} + Bu\\
    &A=E^{-1}A_0, B=E^{-1}B_0, u=\tau\\
    &E = \begin{bmatrix}1&0&0&0\\0&1&0&0\\0&0&a+2b+c&a+b\\0&0&a+b&a\end{bmatrix}\\
    &A_0 = \begin{bmatrix}0&0&1&0\\0&0&0&1\\d&0&0&0\\0&0&0&0\end{bmatrix}, B_0 = \begin{bmatrix}0\\0\\0\\1\end{bmatrix}\\
    &\left\{\begin{array}{l}a=(m_b+m_w)r_w^2+I_w\\b=m_br_wl\\c=m_bl^2+I_b\\d=m_bgl\end{array}\right.
  \end{align}

  The gain $K$ of the feedback control $u=-Kx$ is calculated using the Ricatti equation by applying the optimal regulator to this state equation.
  The transitional direction is controlled by changing $\phi$ to $\phi_{ref}-\phi$.
  The rotational direction $\psi$ is controlled with the proportional control, which adds the reversed torque calculated by multiplying the coefficient to the difference between the current and reference values to the left and right wheels as shown below, assuming that the influence to $\theta, \phi$ is not large.
  \begin{align}
    u_l &= u_l - K_{\psi}(\psi_{ref}-\psi)\\
    u_r &= u_r + K_{\psi}(\psi_{ref}-\psi)
  \end{align}
  $K_{\psi}$ is the proportional gain, $u_{\{l, r\}}$ is each torque of the left and right wheel, and $\psi_{ref}$ is the reference of the rotational direction.

  The state of $\theta, \dot{\theta}, \phi, \dot{\phi}, \psi$ is obtained from the IMU placed on the base and encoders of the wheels.
  In addition, we set $Q = \begin{bmatrix}500.0&1.0&500.0&0.2\end{bmatrix}^T, R = \begin{bmatrix} 0.0001 \end{bmatrix}$ in the evaluation equation $J=\int(\bm{x}^TQ\bm{x} + u^TRu)$ of the optimal regulator, and so, the trackability of $\phi$ is small compared with $\theta$ because the stabilization of the posture is the most important.
}%
{%
  倒立二輪の制御は一般的な状態フィードバックによる方法を用いる.
\figref{figure:pendulum-control}のように, $\theta$をbaseの傾き, $\phi$をbaseからの車輪の回転, $m_w$を車輪の質量, $m_b$をbaseの質量, $I_w$を車輪の慣性モーメント, $I_b$を本体の慣性モーメント, $R$を車輪の半径, $L$を車輪と本体重心間の距離, $\tau$を車輪にかけるトルク, $\bm{x}=\begin{bmatrix}\theta&\phi&\dot{\theta}&\dot{\phi}\end{bmatrix}^T$を状態変数とすると, ラグランジュ方程式を解くことで状態方程式は以下のようになる。
  \begin{align}
    &E\dot{\bm{x}} = A_0\bm{x} + B_0\tau\\
    &\dot{\bm{x}} = A\bm{x} + Bu\\
    &A=E^{-1}A_0, B=E^{-1}B_0, u=\tau\\
    &E = \begin{bmatrix}1&0&0&0\\0&1&0&0\\0&0&a+2b+c&a+b\\0&0&a+b&a\end{bmatrix}\\
    &A_0 = \begin{bmatrix}0&0&1&0\\0&0&0&1\\d&0&0&0\\0&0&0&0\end{bmatrix}, B_0 = \begin{bmatrix}0\\0\\0\\1\end{bmatrix}\\
    &\left\{\begin{array}{l}a=(m_b+m_w)r_w^2+I_w\\b=m_br_wl\\c=m_bl^2+I_b\\d=m_bgl\end{array}\right.
  \end{align}
  この状態方程式に対して$u=-Kx$となるようなフィードバックゲインを最適レギュレータによってリカッチ方程式から求め, 制御している.
  並進方向は$\phi$を指令値との差分$\phi_{ref}-\phi$に変更することで制御し, 回転方向$\psi$に関しては$\theta, \phi$に対する大きな影響はないと考え, 左右の車輪それぞれに逆転したトルクを, 目標に対する差分に対して係数をかけた値を上乗せするP制御を行っている.
  \begin{align}
    u_l &= u_l - K_{\psi}(\psi_{ref}-\psi)\\
    u_r &= u_r + K_{\psi}(\psi_{ref}-\psi)
  \end{align}
  ここで, $K_{\psi}$は比例ゲイン, $u_{\{l, r\}}$は左右それぞれに対するトルク, $\psi_{ref}$は回転方向に対する指令値である.

状態$\theta, \dot{\theta}, \phi, \dot{\phi}, \psi$に関しては体幹についたIMUと車輪のエンコーダ値から求めている.
また, 姿勢の安定化が最も重要であるため, 最適レギュレータの評価式$J=\int(\bm{x}^TQ\bm{x} + u^TRu)$において, $Q = \begin{bmatrix}500.0&1.0&500.0&0.2\end{bmatrix}^T, R = \begin{bmatrix} 0.0001 \end{bmatrix}$としており, $\theta$に対して$\phi$の追従は非常に小さくなっている.
}%

\subsection{Control of TWIMP} \label{subsec:twimp-control}
\switchlanguage%
{%
  The parameters of the state feedback are changed to a great extent due to the motion of the arms.
  Especially, the most important problem is the fluctuation of the center of mass, and it is necessary to deflect the body in order to keep the attitude when the arms are put forward.
  Therefore, the adaptation to the current center of mass is needed by changing $\theta_{ref}$ in response to $\phi$, as shown below:
  \begin{align}
    \theta_{ref} = \theta_{ref} + K_{adapt}\phi \label{eq:adapt}
  \end{align}
  where $K_{adapt}$ is the proportional gain.
  This adaptation control has a dead zone, and $\theta_{ref}$ has to be changed gradually because sudden modification of $\theta_{ref}$ causes the robot to vibrate.
}%
{%
  手が動作することにより, 状態フィードバックのパラメータは大きく異なってくる.
  特に, 手が動くことによる重心位置の変動は最も大きな問題であり, 手を前に出した場合は姿勢を保つために体を反らせる必要が生じる.
  そのため, 以下のように$\theta_{ref}$を$\phi$に応じて変化させて現在の重心状態に適応していく必要がある.
  \begin{align}
    \theta_{ref} = \theta_{ref} + K_{adapt}\phi \label{eq:adapt}
  \end{align}
  ここで, $K_{adapt}$は比例ゲインである.
  また, $\phi$に対しては不感帯を設けており, $\theta_{ref}$は大きく変更し過ぎると振動してしまうため, 徐々に変更していくようになっている.
}%

%%%%%%%%%%%%%%%%%%%%%%%%%%%%%%%%%%%%%%%%%%%%%%%%%%%%%%%%%%%%%%%%%%%%%%%%%%%%%%%%
\section{Preliminary Experiments} \label{sec:experiments}
\switchlanguage%
{%
  In this section, we will first explain the basic movement experiments using the lower and upper limb of TWIMP.
  Next, we will explain simple manipulation experiments using the musculoskeletal dual arm.
  Finally, we will explain several experiments regarding impact resistance.
}%
{%
  まず, TWIMPにおける倒立二輪を用いた移動について実験を行う.
  次に, 筋骨格双腕を用いた簡単なマニピュレーション実験を行う.
  最後に, TWIMPの衝撃対応実験をいくつか行う.
}%

\begin{figure}[t]
  \centering
  \includegraphics[width=0.95\columnwidth]{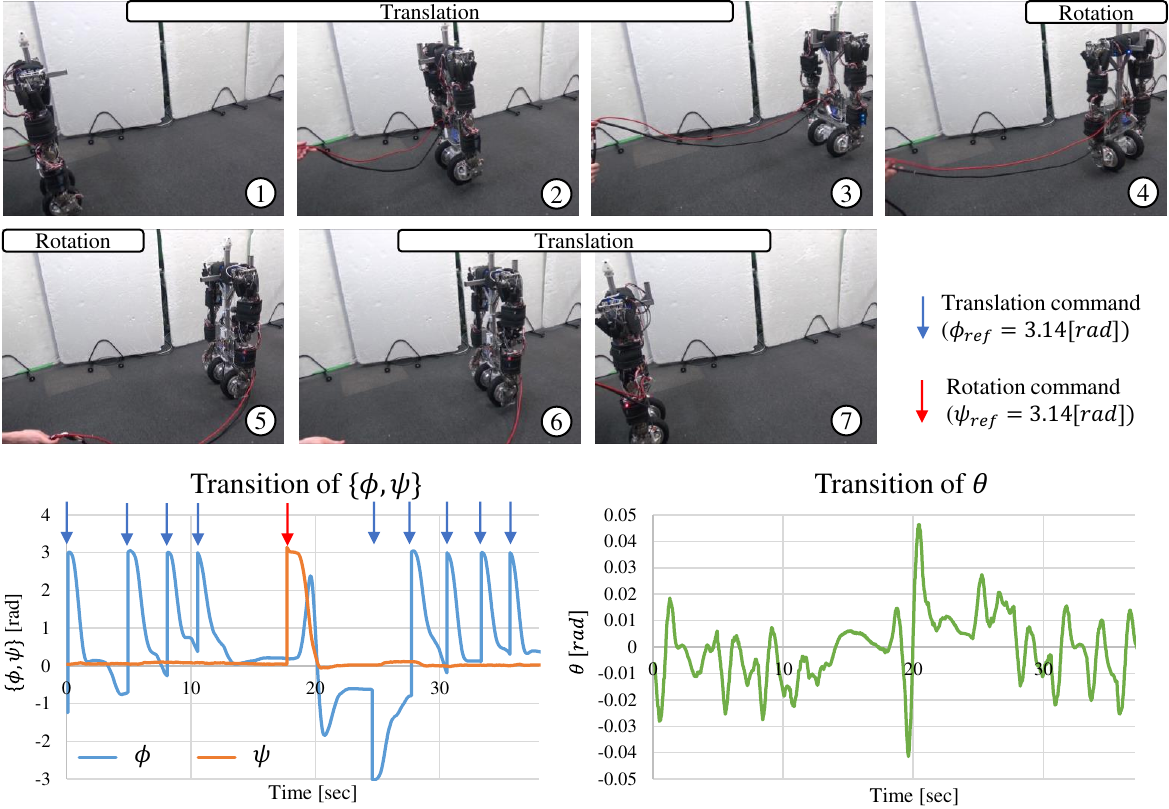}
  \caption{The experiment of translational and rotational movements with the arms fixed to the initial pose.}
  \label{figure:linear-rotate-experiment}
  \vspace{-1.0ex}
\end{figure}

\begin{figure}[t]
  \centering
  \includegraphics[width=0.95\columnwidth]{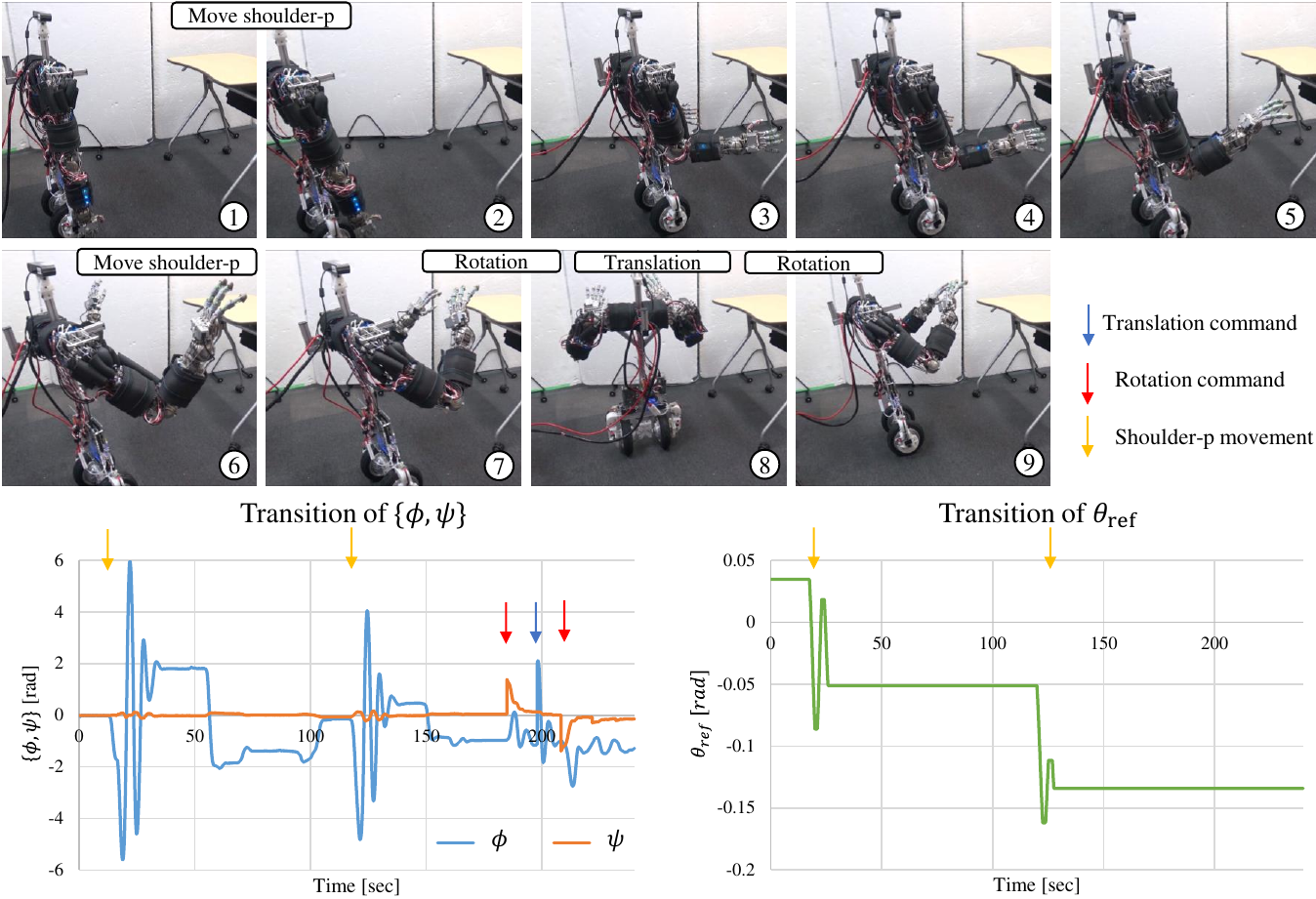}
  \caption{Stabilization of TWIMP while moving its arms}
  \label{figure:upperlimb-movement-experiment}
  \vspace{-1.0ex}
\end{figure}

\subsection{Basic Movements of TWIMP} \label{subsec:basic-movements}
\switchlanguage%
{%
  We conducted an experiment of translational and rotational movements with the arms fixed to the initial pose.
  First, we commanded $\phi_{ref}=3.14[\textrm{rad}]$ several times, making the robot go forward.
  After that, we commanded $\psi_{ref}=3.14[\textrm{rad}]$, making the robot turn around, and then, we commanded $\phi_{ref}=3.14[\textrm{rad}]$ again several times, making the robot go back to the original position.
  The result is shown in \figref{figure:linear-rotate-experiment}.
  The values of $\phi$ and $ \psi$ increased to 3.14 [rad] immediately after the command, and correctly approached zero.
  In addition, the angle of attitude $\phi$ is in the range of $-0.05\sim0.05[\textrm{rad}]$ during all motions.

  Next, we conducted an experiment to stabilize the body while moving the arms.
  The result is shown in \figref{figure:upperlimb-movement-experiment}.
  Due to the large movement in the center of mass when the pitch joints of the shoulders moved, $\phi$ moved greatly but converged correctly by changing $\theta_{ref}$ considering the value of $\phi$ in accordance with \equref{eq:adapt}, though there is some offset.
  In addition, the robot moved stably with its arms raised by converging $\phi$ and $\psi$ when we sent translation and rotation commands.
}%
{%
  まず, 腕を初期状態に固定して並進・回転に関する移動実験を試みる.
  初めに, $\phi_{ref}=3.14[\textrm{rad}]$を何度か指令して並進し, $\psi_{ref}=3.14[\textrm{rad}]$を指令して回転し, もう一度何度か$\phi_{ref}=3.14[\textrm{rad}]$を指令して戻る動作を行う.
  結果を\figref{figure:linear-rotate-experiment}に示す.
  指令を送った直後は$\phi, \psi$の値は3.14[rad]まで上昇し, その後, 0付近まで正しく移動することができている.
  また, 姿勢角$\theta$はどの動作中も$-0.05\sim0.05[\textrm{rad}]$に収まっていることがわかる.

  次に, 上肢を動作させた際に安定を保てるかについて実験を行った.
  結果を\figref{figure:upperlimb-movement-experiment}に示す.
  肩のpitch軸を動かす際は倒立振子の重心位置が大きく変わるため, $\phi$が大きく移動しているが, \equref{eq:adapt}によってその$\phi$のズレから姿勢の指令値$\theta_{ref}$を変更することによって, オフセットは乗っているものの, $\phi$が正しく収束していることがわかる.
  また, 腕を挙げた状態において回転, 並進の指令を送った場合でも, $\phi, \psi$を正しく収束させ, 安定的に動いていることがわかる.
}%

\subsection{Manipulation Experiments} \label{subsec:basic-manipulation}
\begin{figure}[htb]
  \centering
  \includegraphics[width=0.95\columnwidth]{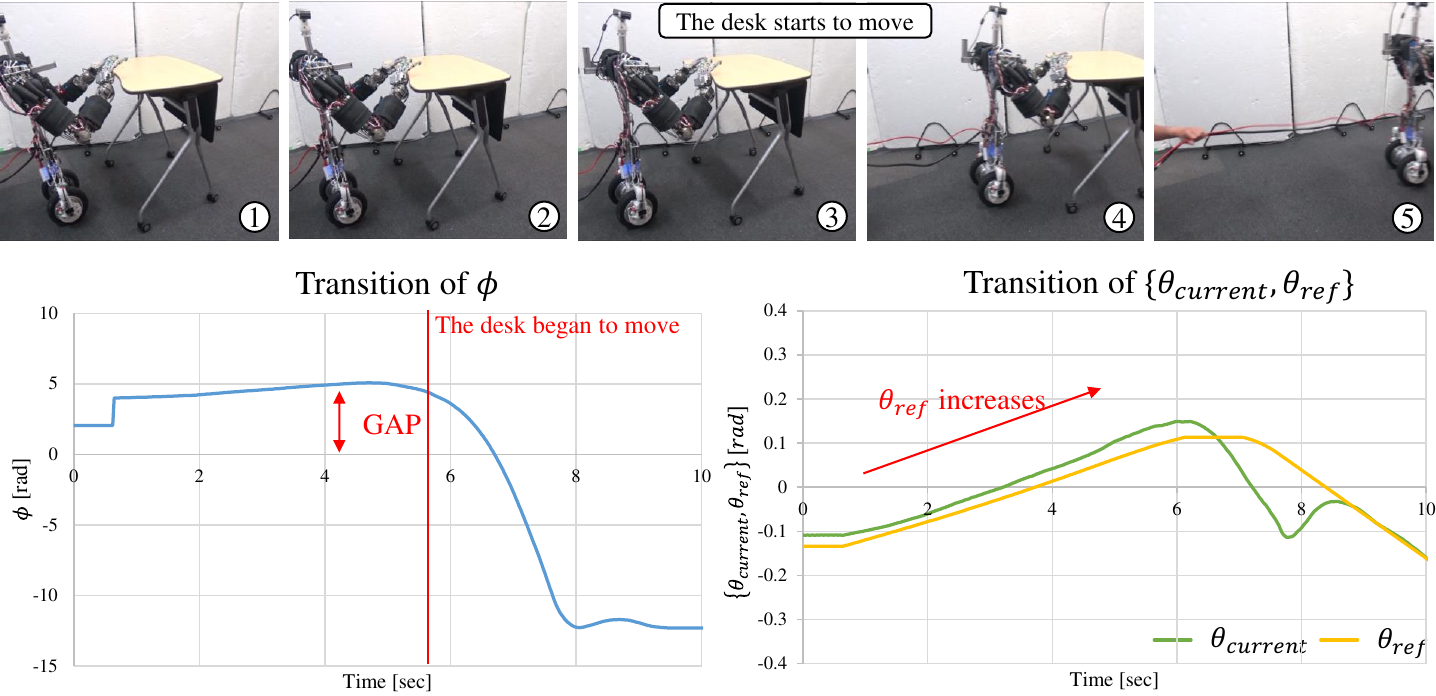}
  \caption{The experiment of pushing a movable desk with casters.}
  \label{figure:table-push-experiment}
  \vspace{-1.0ex}
\end{figure}

\begin{figure*}[htb]
  \centering
  \includegraphics[width=1.95\columnwidth]{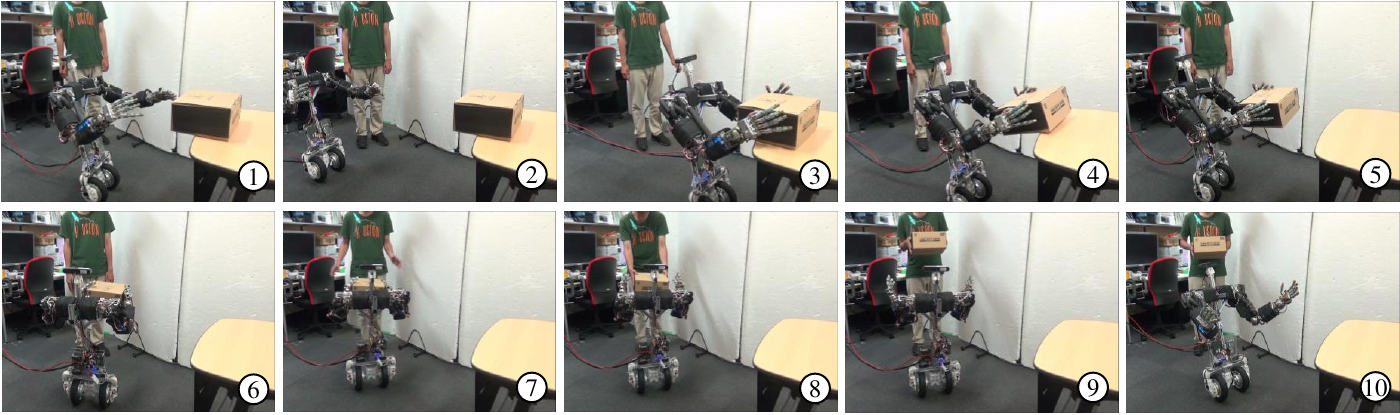}
  \caption{The experiment of handing a box with both arms.}
  \label{figure:box-manipulation-experiment}
  \vspace{-1.0ex}
\end{figure*}

\switchlanguage%
{%
  First, we conducted the experiment of pushing a movable desk with casters.
  The arms were raised to the height of the desk.
  We commanded $\phi_{ref}$ to the front, and the attitude of the robot gradually tilted and pushed the desk by \equref{eq:adapt}.
  The result is shown in \figref{figure:table-push-experiment}.
  $\theta_{ref}$ is gradually changed and the robot succeeded in pushing the desk.

  Next, we conducted the experiment of holding a box with both hands and handing it to a person.
  The robot held a box placed on a desk with both arms, turned to a person, and released it when he held it.
  The result is shown in \figref{figure:box-manipulation-experiment}.
  The robot succeeded in holding a box by pressing from both sides and in carrying the box to a person by commanding $\phi_{ref}$ and $\psi_{ref}$.
}%
{%
  まず, 腕と倒立二輪を用いてキャスター付き机を押す実験を行う.
  双腕を机が押せるような位置まで動作させ, $\phi_{ref}$を前方に指定することで倒立振子の姿勢が徐々に傾き, 力を伝わらせることで机を押していく.
  その結果を\figref{figure:table-push-experiment}に示す.
  $\theta_{ref}$が徐々に傾き, 最終的に机を押すことに成功しているのがわかる.

  次に, ダンボール箱を双腕によって把持し, それを人に受け渡す実験を行う.
  まず机の上にあるダンボール箱を双腕によって把持し, 回転して人の方向を向き人が箱を持ったところで手を話す.
  結果を\figref{figure:box-manipulation-experiment}に示す.
  筋骨格ヒューマノイド特有の柔らかな腕によってダンボールを横から押さえることによって把持し, $\phi_{ref}, \psi_{ref}$を指令することによって人の場所まで持って行くことに成功している.
}%

\subsection{Stability from External Force} \label{subsec:basic-stability}
\begin{figure}[t]
  \centering
  \includegraphics[width=0.95\columnwidth]{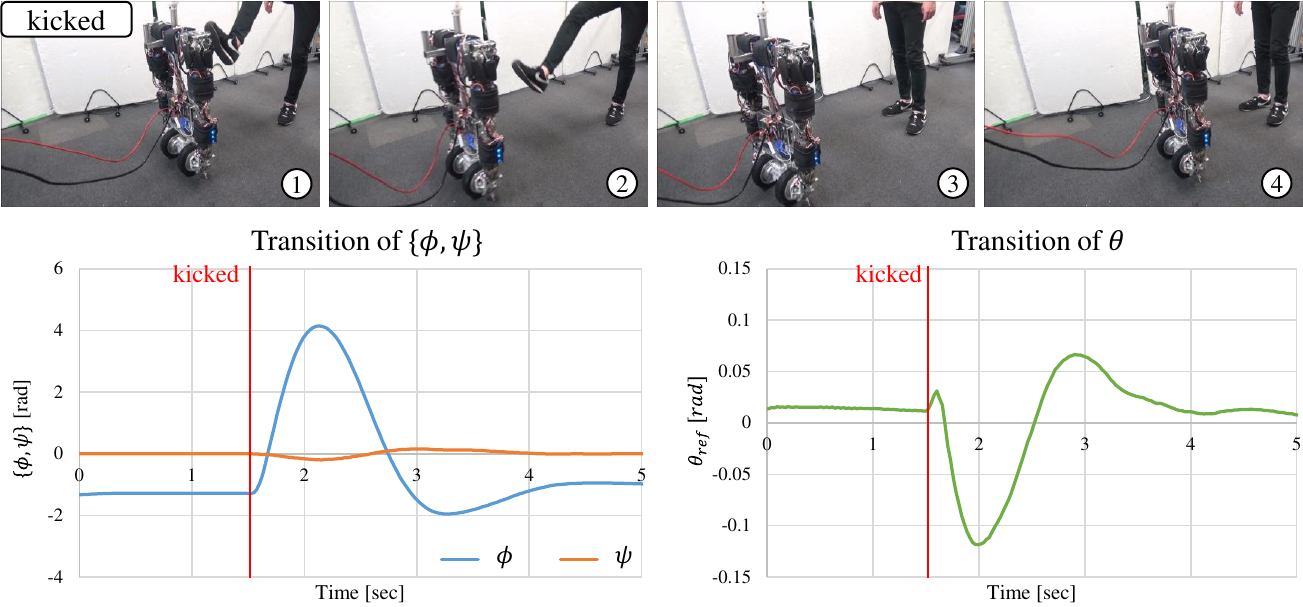}
  \caption{The stability of TWIMP at the initial pose.}
  \label{figure:kick-experiment}
  \vspace{-1.0ex}
\end{figure}

\begin{figure}[htb]
  \centering
  \includegraphics[width=0.95\columnwidth]{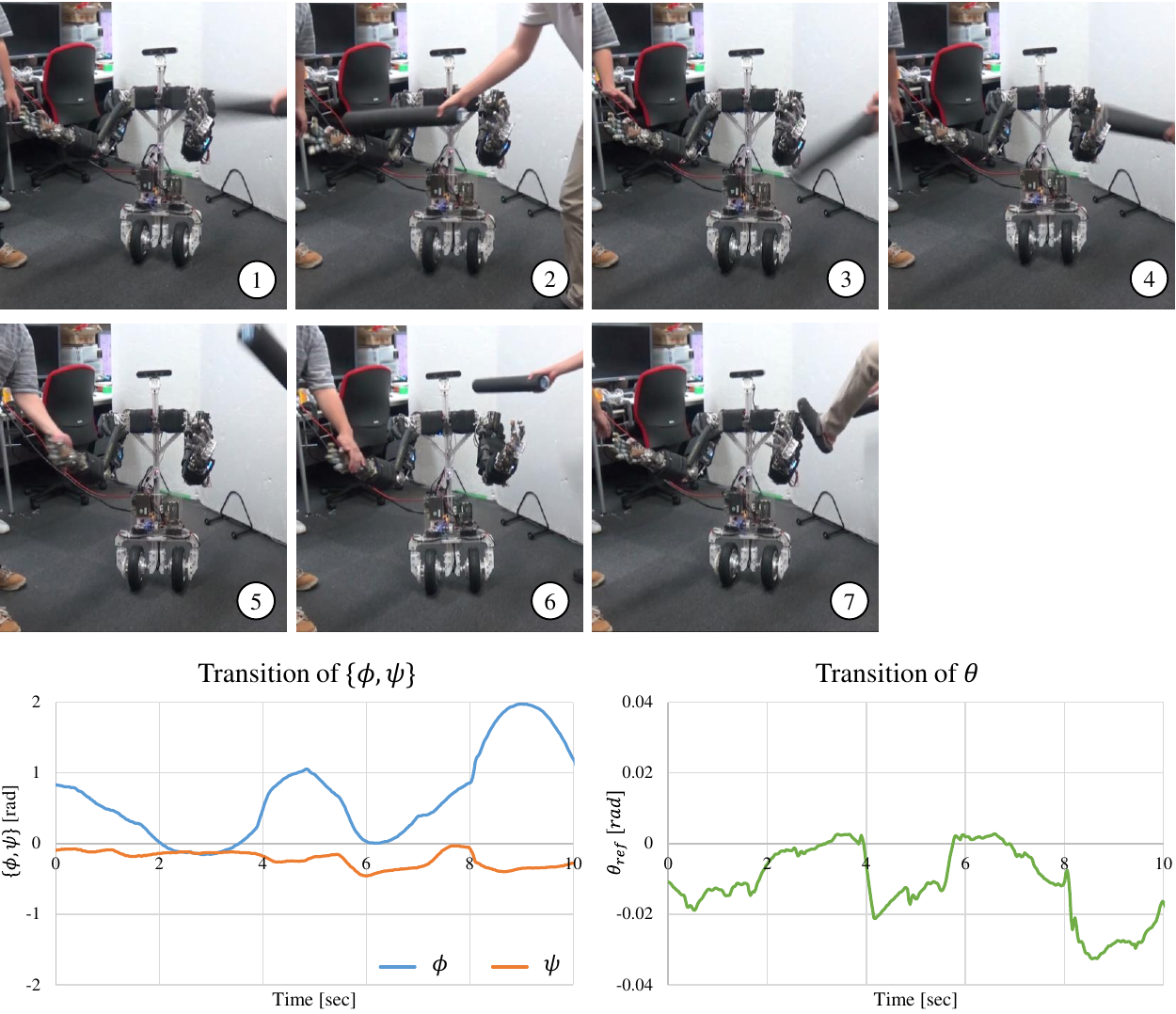}
  \caption{The stability against the impact to the upper limbs.}
  \label{figure:hitting-experiment}
  \vspace{-3.0ex}
\end{figure}

\begin{figure*}[t]
  \centering
  \includegraphics[width=1.85\columnwidth]{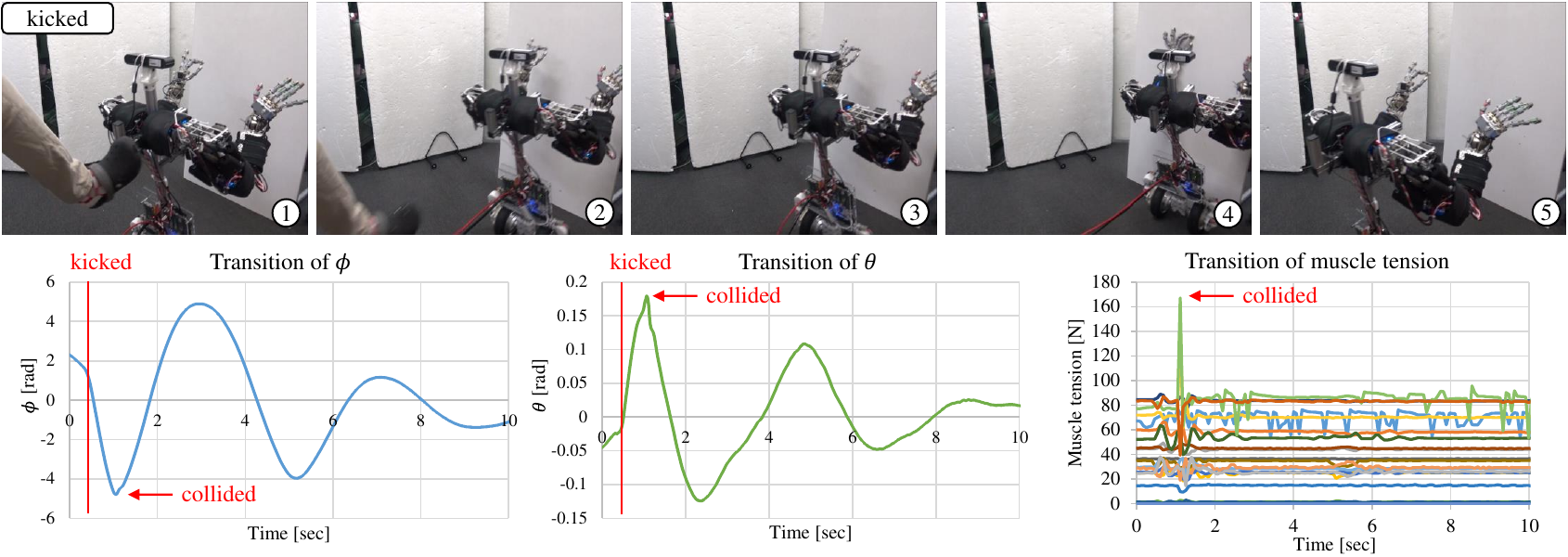}
  \caption{The evaluation of the impact resistance in the experiment of collision with a wall.}
  \label{figure:attack-experiment}
  \vspace{-3.0ex}
\end{figure*}

\switchlanguage%
{%
  We investigated the stability against an impact by kicking the robot when the arms are in the initial position.
  The result is shown in \figref{figure:kick-experiment}.
  The values of $\phi$ and $\psi$ are converged in about 4 sec, though they are changed significantly at the time when we kicked the center of the robot.

  Next, we investigated the stability with the arms raised when we hit the arms and base links.
  The result is shown in \figref{figure:hitting-experiment}.
  The robot was able to stabilize the attitude though $\phi$, $\psi$, and $\theta$ were affected when we added short and long impact.

  Finally, we investigated the resistance ability when an impact is added and the robot collides with a wall with the arms raised.
  The result is shown in \figref{figure:attack-experiment}.
  $\phi$ and $ \theta$ were changed at the moment of impact and changed rapidly when the robot collided with the wall.
  In addition, the muscle tension of the arms increased to more than twice the original value.
  However, the soft body structure absorbed the shock, and the tension and attitude returned to their original values.
}%
{%
  まず, 手が初期状態であるときに, 体を蹴ることで衝撃に対する安定性を見る.
  結果を\figref{figure:kick-experiment}に示す.
  体の中心を蹴った瞬間, 大きく$\phi, \theta$が変動しているが, 4[sec]程度で収束していることがわかる.

  次に, 手を前に出した状態において手や倒立振子を叩いた場合における安定性を見る.
  結果を\figref{figure:hitting-experiment}に示す.
  手に時間的に短い衝撃と時間的に長い衝撃等を繰り返し与えた際, $\phi, \psi, \theta$どれもがその影響を受けてはいるが, 安定して姿勢を保っていることがわかる.

  最後に, 手を前に出した状態において上体に衝撃を加え, 壁に衝突した際の対応について見てみる.
  結果を\figref{figure:attack-experiment}に示す.
  蹴られた瞬間に$\phi, \theta$が動き出し, 壁に衝突した瞬間に$\phi, \theta$が急激に変化している.
  また, 腕の筋張力も倍以上に跳ね上がっていることがわかる.
  しかし, 柔らかい体により衝撃を吸収して筋張力は元に戻り, 姿勢も安定に戻っていくことがわかる.
}%

\section{Discussion} \label{sec:discussion}
\switchlanguage%
{%
  We will start by considering the results of the conducted experiments.

  First, while the robot moves stably in the transition and rotation experiment of \secref{subsec:basic-movements}, we can see that the posture $\theta$ is changed greatly especially with the rotation of $\psi$.
  It seems that this is due to the fact that the rotation of $\psi$ is not taken into account when considering the model of the state equation, though it is considered in several controllers such as \cite{li2008controller}.
  Also, since we had to limit the speed of TWIMP, it is necessary to consider the translation ability and stability of the posture under a more dynamic system.

  Second, during the movements of the arms in \secref{subsec:basic-movements}, $\phi$ vibrates greatly at first and settles down in the end.
  This behavior of the robot will be improved by calculating the track of the center of mass using the model of the upper limb and feed forward control of $\bm{\theta}_{ref}$ in addition to the control scheme of \secref{subsec:twimp-control}.
  However, we cannot decide if this is the best method, since the musculoskeletal structure is soft and there is a large error between the actual robot and its geometric model.
  In addition, we can consider installing a different control system such as the adaptive control.
  This discussion also occurs when considering the pushing desk motion in \secref{subsec:basic-manipulation}, and we have to rethink the whole control scheme of this robot to manipulate objects stably.

  Third, we investigated stability resistance of TWIMP from impact in \secref{subsec:basic-stability}, and these results seem to be due to the mobility and stability of the two-wheel inverted pendulum and the soft body structure of the musculoskeletal humanoid.
  However, although this robot has a high impact resistance ability and is not damaged when falling down, it has not succeeded in getting up yet.

  Next, we will state future works of this study.
  We have discussed the problems of the present control methods, and we should examine motion generation using learning control methods.
  Learning control methods seem to be effective in not only solving the problems of modeling the soft body of the musculoskeletal humanoid, but also in moving in the real world with environmental contact.
  Since the musculoskeletal upper limb is soft and has tolerance to impact, it is rarely damaged when falling down and is suitable for motion learning in the real world.
  We would like to achieve advanced tasks such as livelihood support and care service.
  In addition, we should rethink the problems of robot hardware.
  Since TWIMP has to compensate for the change in the center of mass due to the arm movement by using only the base posture alteration, the whole robot posture changes according to the movement of the arms.
  We can solve the problem by adding either a spine or tail.
  The robot can change its posture freely and move more flexibly by modifying the trunk link to resemble the flexible spine structure, extending the design of the musculoskeletal upper limb.
  In addition, we suppose that a tail link will enable the stabilization of the robot body posture by changing the tail posture according to the arm movement, or by using the tail as a stand to support the body.
  As shown in this section, we believe that the two-wheel inverted musculoskeletal pendulum and its advancements will become the first step in the development of the learning robot, which is useful in the real world thanks to its soft hardware structure and mobility.
}%
{%
  初めに, ここまで行ってきた実験について考察を行う.

  1つ目に, \secref{subsec:basic-movements}の前半における移動実験は安定的に動作してはいるものの, 特に$\psi$の回転を伴う場合は, 姿勢$\theta$の変動が大きくなっていることがわかる.
  これは, 状態フィードバックのモデルとして回転方向のモデルが入っていないからであると思われる.
  また, スピードは現状抑えており, より動的なシステムになった際の移動性能, 安定性能についても考える必要がある.

  2つ目に, \secref{subsec:basic-movements}の後半における上肢の動作を行う際は, $\phi$が大きくズレて振動しながら最終的に収束していることがわかる.
  これは, \secref{subsec:twimp-control}における制御において, 重心の移動を上肢のモデルから導出し, $\bm{\theta}_{ref}$をfeedforwardで与えることでより改善されると考えられる.
  しかし, 筋骨格ヒューマノイドの身体は柔らかく, モデルと現実が合いにくいという問題もあるためこれが一概に良いとは言えない.
  また, 適応制御のような枠組みを用いることも考えられる.
  これは\secref{subsec:basic-manipulation}における机を押す動作も同じであり, より安定的に物体をマニピュレーションするためには制御を考えなおす必要がある.

  3つ目, \secref{subsec:basic-stability}においては, 衝撃に対する安定性を評価しているが, これは倒立振子の非常に優れた移動性能と安定性能, 筋骨格ヒューマノイドの柔軟な身体構造の利点がよく活かされていると考える.
  しかし, 現状倒れても壊れない等の衝撃吸収性能を有しているものの, 単純な制御では起き上がり動作には成功していない.

  次に, 今後の展望について述べる.
  先に現状の制御の問題点を議論したが, 今後はより学習的な手法による動作生成を考えるべきだと感じている.
  筋骨格ヒューマノイドの柔軟な身体がモデル化困難なだけでなく, 環境接触を伴う場合はより学習的な手法が効いてくると思われる.
  筋骨格型上肢は柔軟であり衝撃に強く, 転倒しても壊れることは殆ど無く, 実世界における学習にも適している.
  そして, より上位の日常生活における支援や介護支援等も行っていきたい.
  また, ハードウェアに関しても問題点があると考える.
  現状腕を動かした場合はその重心の変化をbaseの姿勢のみで担保しなければならず, 腕を動かすと全体の姿勢が変わってしまう.
  そこで, 体幹リンクをつけるという方法と, カウンターウェイトとしての尻尾をつけるという二つの方法が考えられる.
  体幹を筋骨格型の設計思想を延長して背骨のような構造にすることで, 姿勢を自在に変化させ, より柔軟な動きができる可能性がある.
  また, 尻尾をつけることで, udeの姿勢に応じて尻尾の姿勢を変えて常に姿勢を一定に保つことができ, 尻尾を地面につけることによって, 力を全く発揮せずに姿勢を安定化させることもできると考えられる.

  これらのように, 本研究で開発した筋骨格型倒立二輪とその発展系が, 柔軟な構造を持ち, かつ移動性能に優れた, 実世界で役に立つ学習型ヒューマノイドの形の第一歩となると確信している.
}%

%%%%%%%%%%%%%%%%%%%%%%%%%%%%%%%%%%%%%%%%%%%%%%%%%%%%%%%%%%%%%%%%%%%%%%%%%%%%%%%%
\section{CONCLUSION} \label{sec:conclusion}
\switchlanguage%
{%
  In this research, we proposed a two wheel inverted musculoskeletal pendulum: TWIMP, which is composed of a musculoskeletal upper limb and two-wheel inverted pendulum, as one of the platforms for researching learning control methods with environmental physical interactions in the real world.
  This robot has two advantages, which are the ability to contact the environment softly using the flexible structure of the musculoskeletal upper limb and variable stiffness mechanism, and the mobility of the two-wheel inverted pendulum which has a small footprint.
  We conducted several experiments concerning the ability of locomotion, manipulation, and durability using TWIMP, and we were able to show the potential ability by the combination of the soft upper body and the movable lower body.
  In future work, we would like to study the learning control methods and advanced hardware design of TWIMP to make it useful in several situations such as livelihood support and care service.
}%
{%
  本研究では, 環境接触を伴う学習型制御開発のプラットフォームの一つとして, 筋骨格上肢と倒立二輪を組み合わせた筋骨格倒立二輪ヒューマノイドTWIMPを提案した.
  筋骨格上肢の柔軟な外界との接触や可変剛性機構等の利点と, 倒立二輪のフットプリントが小さく走破性能に優れる利点を組み合わせた構成である.
  この筋骨格倒立二輪TWIMPを用いて, 移動性能・マニピュレーション性能・対衝撃性能に関していくつかの実験を行い, 柔軟な身体と移動性能を組み合わせることのポテンシャルを示すことができた.

  今後は, このTWIMPを使って, 学習的に起き上がる動作や移動しながらのマニピュレーション動作等の生成, また, 日常生活支援や災害対応等, 様々なシーンにおいて活躍できるロボットを目指して制御手法・ハードウェア変更の模索をしていきたい.
}%

\section*{Acknowledgement}
The authors would like to thank Yuka Moriya (Ochanomizu University) for proofreading this manuscript.

{
  %\footnotesize
  %\small
  %\bibliographystyle{junsrt}
  \bibliographystyle{IEEEtran}
  \bibliography{main}
}

\end{document}